\documentclass{article} 
\usepackage{iclr2026_conference, times}
\usepackage{graphicx}  
\usepackage{fontawesome5} 

\usepackage{amsmath,amsfonts,bm}

\def\eqref#1{equation~\ref{#1}}

\def\1{\bm{1}}

\DeclareMathAlphabet{\mathsfit}{\encodingdefault}{\sfdefault}{m}{sl}
\SetMathAlphabet{\mathsfit}{bold}{\encodingdefault}{\sfdefault}{bx}{n}

\DeclareMathOperator*{\argmax}{arg\,max}

\usepackage{pifont}

\usepackage{hyperref}
\usepackage{url}
\usepackage{subfigure}
\usepackage{wrapfig}
\usepackage[utf8]{inputenc} 
\usepackage[T1]{fontenc}    
\usepackage{hyperref}       
\usepackage{url}            
\usepackage{booktabs}       
\usepackage{amsfonts}       
\usepackage{nicefrac}       
\usepackage{microtype}      
\usepackage{xcolor}         
\usepackage{tikz}

\usepackage{booktabs}
\usepackage{subfigure}
\usepackage{enumitem}
\usepackage{rotating}
\usepackage[utf8]{inputenc}
\usepackage{amsmath,amssymb,amsfonts}

\usepackage{booktabs,bm}
\usepackage{colortbl}
\usepackage{multirow}
\usepackage{tabularx}
\usepackage{amsmath}       
\usepackage{amsfonts}      

\usepackage{xcolor}
\definecolor{commentcolor}{rgb}{0.5, 0.5, 0.5} 
\usepackage{booktabs}
\usepackage{adjustbox}
\definecolor{grey}{rgb}{0.89,0.71,0.57}
\definecolor{pink}{rgb}{1,0.94,1}
\definecolor{purple}{rgb}{0.84,0.78,1}
\definecolor{white}{rgb}{1,1,1}
\usepackage{amsmath}
\usepackage{mathtools}
\usepackage{needspace}
\usepackage[utf8]{inputenc}
\usepackage{hyperref}
\usepackage{comment}
\usepackage{algorithm}
\usepackage{algorithmic}
\usepackage{wrapfig}
\usepackage{amsmath}
\usepackage{graphicx}
\usepackage{subcaption}
\usepackage{multirow}
\usepackage{booktabs}
\definecolor{Nuwa_blue}{rgb}{0,205,205}

\definecolor{rowblue}{RGB}{230, 240, 255}
\definecolor{rowgreen}{RGB}{230, 255, 230}
\definecolor{rowyellow}{RGB}{255, 255, 224}
\definecolor{rowpink}{RGB}{255, 230, 240}
\definecolor{rowcyan}{RGB}{224, 255, 255}
\definecolor{roworange}{RGB}{255, 240, 224}
\definecolor{rowviolet}{RGB}{240, 230, 255}
\definecolor{rowlime}{RGB}{240, 255, 224}
\definecolor{rowmint}{RGB}{224, 255, 240}
\definecolor{rowapricot}{RGB}{255, 224, 224}
\definecolor{rowgray}{RGB}{235, 235, 235}

\definecolor{TAMLink}{HTML}{245F69}
\hypersetup{
    colorlinks=true,
    linkcolor=TAMLink,
    citecolor=TAMLink,
    filecolor=TAMLink,
    urlcolor=TAMLink,
    }
\usepackage[utf8]{inputenc}
\usepackage{amsmath}
\usepackage{amssymb}
\usepackage{enumitem}
\usepackage{xcolor}
\usepackage{graphicx} 
\usepackage{tcolorbox} 

\newlist{researchquestions}{enumerate}{1}
\setlist[researchquestions]{
    label=\textbf{RQ\arabic*:}, 
    leftmargin=*, 
    itemsep=8pt,
    font=\normalfont
}

\newcommand{\method}{\textsc{TAM}}
\newcommand{\methodfull}{Task-Aware Memory Distillation}

\title{TAM: Task-Aware Memory Distillation\\for Efficient Spatiotemporal Prediction}

\iclrfinalcopy
\author{
  \href{https://openreview.net/profile?id=~Yuqi_Li7}{\textbf{\textcolor{black}{Yuqi Li}}}$^{1}$ \quad
  \href{https://openreview.net/profile?id=~Xiaoqin_Feng2}{\textbf{\textcolor{black}{Xiaoqin Feng}}}$^{2}$ \quad
  \href{https://openreview.net/profile?id=~Fan_Xu5}{\textbf{\textcolor{black}{Fan Xu}}}$^{3}$ \quad
  \href{https://openreview.net/profile?id=~Weilun_Feng2}{\textbf{\textcolor{black}{Weilun Feng}}}$^{4}$ \quad
  \href{https://openreview.net/profile?id=~Chuanguang_Yang1}{\textbf{\textcolor{black}{Chuanguang Yang}}}$^{4}$ \\
  \href{https://openreview.net/profile?id=~Yingli_Tian1}{\textbf{\textcolor{black}{Yingli Tian}}}$^{1}$ \quad
  \href{https://openreview.net/profile?id=~Hao_Wu39}{\textbf{\textcolor{black}{Hao Wu}}}$^{5,*}$ \\[1ex]
  {\normalfont\small $^{1}$CUNY City College of NY} \\
  {\normalfont\small $^{2}$Wyze Labs, Inc.} \\
  {\normalfont\small $^{3}$University of Science and Technology of China} \\
  {\normalfont\small $^{4}$Institute of Computing Technology, Chinese Academy of Sciences} \\
  {\normalfont\small $^{5}$Tsinghua University} \\[0.5ex]
  {\normalfont\small $^{*}$Corresponding author}
}

\usepackage{newtxmath}

\DeclareCaptionFont{tamgaramond}{\fontfamily{EBGaramond-LF}\selectfont}

\begin{document}

\maketitle

\begin{abstract}
Knowledge distillation enables efficient spatiotemporal prediction by transferring knowledge from an accurate teacher to a compact student. However, matching outputs or features independently for each sample leaves cross-sample predictive structure underused. Exploiting this structure requires representations and historical references that reflect the dynamics of each task. We propose \method{}, a \methodfull{} framework that organizes a frozen teacher's knowledge into a bounded, retrievable history. Memory entries encode latent features, forecast changes, or flow residuals, while task-specific selection rules identify relevant historical references. The student either matches the teacher's similarity distribution over shared references or regresses observation-conditioned residual prototypes. These objectives complement supervised prediction and conventional distillation. The teacher, memory, and auxiliary adapters are used only during training, leaving student inference unchanged. We evaluate \method{} on video prediction, weather forecasting, and traffic flow prediction across multiple teacher--student configurations. Averaged over four paired runs, adding \method{} improves SSIM on all six video datasets and reduces MSE on five relative to the corresponding KD baselines. Mean paired MSE reductions reach $1.86\%$ on KittiCaltech, $1.93\%$ on WeatherBench with a gSTA teacher, and $1.01\%$ on TaxiBJ. These results demonstrate the utility of historical teacher supervision across distinct forecasting tasks without additional student inference cost.
\end{abstract}

\section{Introduction}
\label{sec:introduction}

Spatiotemporal prediction supports robotic interaction~\citep{finn2016unsupervised,ebert2018visualforesight}, weather forecasting~\citep{shi2015convlstm,ravuri2021skilful,gao2022earthformer}, and traffic modeling~\citep{zhang2017stresnet}. Recurrent models capture spatiotemporal states~\citep{wang2017predrnn}, non-stationary dynamics~\citep{wang2019mim}, and motion through historical-state recall~\citep{wang2019eidetic,chang2021mau}. Convolutional, attention-based, and physics-guided predictors offer complementary architectures~\citep{gao2022simvp,tan2023tau,leguen2020phydnet,wu2024earthfarsser,tan2023openstl}. Their advances motivate transferring predictive knowledge to compact students for efficient deployment.

Knowledge distillation provides teacher supervision~\citep{hinton2015distilling} through intermediate hints, attention maps, activation boundaries, and cross-stage features~\citep{romero2015fitnets,zagoruyko2017attention,heo2019activation,chen2021reviewkd}. Structured and channel-wise distillation preserve spatial dependencies in dense prediction~\citep{liu2019structured,shu2021channel}; forecasting methods also transfer frequency-related knowledge and semantic priors~\citep{li2025frequency,wang2026s2kd}. Teacher accuracy alone does not guarantee effective transfer when teacher and student capacities differ~\citep{cho2019efficacy}. These findings motivate selecting supervision that the student can exploit.

A second choice concerns the examples defining supervision. Relational and similarity-preserving distillation transfer dependencies among representations~\citep{park2019relational,tung2019similarity}, while contrastive distillation transfers their discriminative structure~\citep{tian2020crd}. Instance-discrimination memory banks, momentum queues, and cross-batch memory retain references beyond the current mini-batch~\citep{wu2018instance,he2020moco,wang2020xbm}. CIRKD uses historical teacher embeddings for cross-image relational distillation~\citep{yang2022cirkd}. Unlike temporal state memory within a predictor, these references supply cross-sample training supervision. Forecasting therefore raises a task-specific question: which predictive quantities and historical examples should supervise the current dynamics?

\begin{wrapfigure}{R}{0.54\textwidth}
    \vspace{-6pt}
    \centering
    \includegraphics[width=\linewidth]{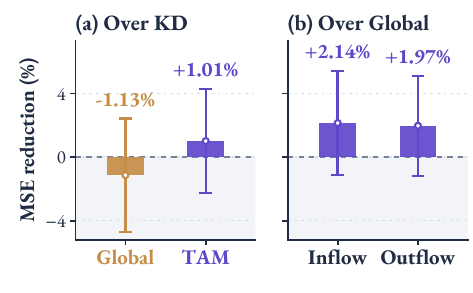}
    \captionsetup{font=footnotesize,skip=3pt}
    \caption{TaxiBJ memory comparison. (a) Overall MSE reduction over KD. (b) TAM's inflow/outflow gains over Global Memory. Bars: four-run paired means; whiskers: population SD. Positive values indicate improvement.}
    \label{fig:tam_pilot}
    \vspace{-7pt}
\end{wrapfigure}

Historical references have different meanings across forecasting tasks. Video dynamics concentrate in moving or occupied regions; weather states depend on geographic location; and future traffic changes depend on the observed flow and its recent trend. A generic feature memory does not explicitly account for these distinctions. We therefore consider two coupled choices: which predictive quantity to store and which historical entries should supervise the current sequence.

Figure~\ref{fig:tam_pilot} compares memory designs on \mbox{TaxiBJ} with a fixed TAU teacher, U-Net-Base student, and output-plus-feature MSE distillation baseline. Both memory variants use matched training settings, 2,048 historical entries, and 16 neighbors. Relative to the no-memory KD baseline, global residual memory increases test MSE by $1.13\%$, whereas \method{} reduces it by $1.01\%$ on average. Channel-wise analysis further shows that \method{} reduces inflow and outflow MSE by $2.14\%$ and $1.97\%$, respectively, relative to global memory. All percentages average relative changes over four paired runs. These results support organizing historical supervision around observation conditions and prediction horizons. However, gains vary across runs, and this joint comparison does not isolate the individual contributions of conditioned retrieval and horizon grouping.

We introduce \method{}, a \methodfull{} framework that makes these choices explicit within a common teacher--student training procedure. A bounded memory stores latent features, forecast changes, or flow residuals from a frozen teacher. Task structure determines the current representation and the eligible historical references. Relational variants match teacher and student similarity distributions over shared references. For traffic flow prediction, observation-conditioned retrieval instead aggregates historical teacher residuals into regression targets for the student.

Memory supervision complements ground-truth prediction, output distillation, and optional feature alignment. The teacher and memory operate only during training, so inference retains the original student architecture and cost. We evaluate the additional effect of memory against matched distillation baselines on video, weather, and traffic prediction.

Our contributions are threefold:
\begin{itemize}[leftmargin=*,itemsep=2pt,topsep=3pt,parsep=0pt,partopsep=0pt]
    \item We formulate task-aware teacher memory as an interface for cross-sample supervision in spatiotemporal prediction, making the stored representation and reference selection explicit.
    \item We develop shared-reference relation matching and observation-conditioned residual prototypes as task-specific memory objectives that complement prediction and distillation.
    \item We evaluate teacher--student combinations and memory designs across three prediction domains, using matched comparisons across repeated runs to characterize the effect of historical supervision.
\end{itemize}

\section{Related Work}
\label{sec:related_work}

\paragraph{Spatiotemporal predictive learning.}
Early recurrent predictors couple spatial processing with temporal state updates. ConvLSTM replaces fully connected transitions with convolutions~\citep{shi2015convlstm}, while PredRNN propagates spatiotemporal memory across both time and network depth~\citep{wang2017predrnn}. PhyDNet separates a dynamics component constrained by partial differential equations from complementary visual information~\citep{leguen2020phydnet}. Recurrent-free models offer another approach: SimVP uses a convolutional encoder--translator--decoder architecture~\citep{gao2022simvp}, and TAU combines spatial and temporal attention with a regularizer on frame differences~\citep{tan2023tau}. Earthformer organizes space--time attention into cuboids for Earth-system forecasting~\citep{gao2022earthformer}, while Earthfarsser studies a shared architecture for different dynamical systems~\citep{wu2024earthfarsser}. OpenSTL provides a common implementation and evaluation framework for recurrent and recurrent-free predictors~\citep{tan2023openstl}. Whereas these approaches develop predictive architectures and objectives, \method{} augments the supervision of an existing student with task-specific teacher history, without modifying its inference architecture.

\paragraph{Knowledge distillation for efficient prediction.}
Knowledge distillation transfers information from a trained teacher to a student~\citep{hinton2015distilling}. FitNets uses intermediate hints~\citep{romero2015fitnets}; attention transfer aligns attention maps~\citep{zagoruyko2017attention}; activation-boundary distillation transfers the partition induced by hidden neurons~\citep{heo2019activation}. ReviewKD studies connections across feature stages~\citep{chen2021reviewkd}, and channel-wise distillation aligns spatial probability maps within channels for dense prediction~\citep{shu2021channel}. Recent forecasting methods further adapt the transferred representations to temporal prediction. Frequency-Aligned KD transfers frequency-related representations~\citep{li2025frequency}, and S$^2$-KD combines semantic priors from a privileged multimodal teacher with spectral representations~\citep{wang2026s2kd}. \method{} complements these objectives by organizing supervision across training sequences. A frozen forecasting teacher supplies historical features or forecast changes, and task-specific rules select references for the current sample. Matched output and feature baselines expose the additional effect of historical supervision.

\paragraph{Relational transfer and historical memory.}
Relational KD transfers distances and angles among examples~\citep{park2019relational}, and similarity-preserving KD maintains pairwise activation similarities~\citep{tung2019similarity}. Contrastive representation distillation uses a contrastive objective to transfer representational information~\citep{tian2020crd}. More broadly, MoCo demonstrates how a queue can maintain a reference dictionary beyond the current mini-batch, with a momentum encoder controlling representation drift~\citep{he2020moco}. The closest predecessor to our relational objective is CIRKD~\citep{yang2022cirkd}: it stores embeddings from a frozen segmentation teacher and aligns pixel-to-pixel and pixel-to-region similarity distributions across images. Our relational objective builds on this shared-reference formulation. We adapt the construction of teacher history to forecasting: motion or occupancy selects video regions, geographic location constrains weather references, and observed traffic flow conditions residual retrieval. The traffic variant further replaces relation matching with residual-prototype regression. Thus, \method{} extends historical supervision through task-specific representations and reference organization, rather than introducing a new queue mechanism or relational divergence.

\section{Method}
\label{sec:method}

\begin{figure}[t]
    \centering
    \includegraphics[width=\linewidth]{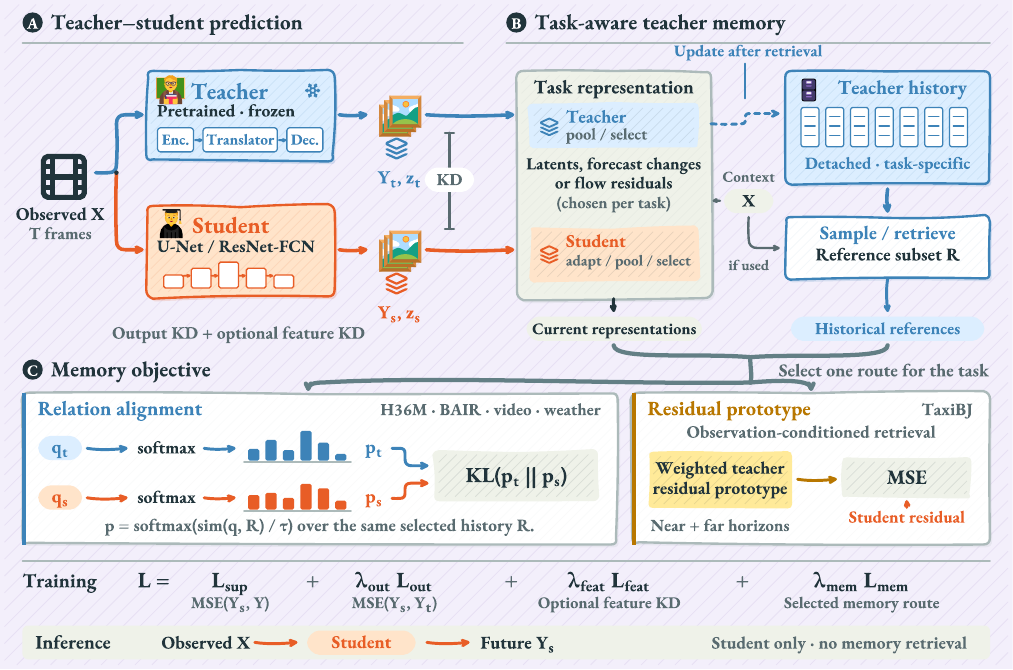}
    \caption{Overview of \method{}. \textbf{(A)} A frozen teacher provides output and optional feature supervision. \textbf{(B)} Task-specific representations enter history after reference selection. \textbf{(C)} Memory supervision matches relations over shared references or regresses retrieved residual prototypes. Inference uses only the student. Glyphs and distribution bars are schematic.}
    \label{fig:method}
\end{figure}

\method{} augments per-sample distillation with supervision from a bounded history of teacher predictions and representations. The task determines both the quantity stored in memory and the historical entries used as references. The student then learns either the teacher's relations to these references or a residual target aggregated from them. Figure~\ref{fig:method} summarizes the framework.

\subsection{Problem Formulation}
\label{sec:tam_setup}

Given observations $X=(X_1,\ldots,X_T)$, spatiotemporal prediction estimates the next $K$ frames or fields, $Y=(Y_1,\ldots,Y_K)$. For a training set $\mathcal D=\{(X^{(n)},Y^{(n)})\}_{n=1}^{N}$, we formulate learning as conditional maximum likelihood:
\begin{equation}
    \theta^\star=\argmax_{\theta}\;\frac{1}{N}\sum_{n=1}^{N}
    \log p_{\theta}\!\left(Y^{(n)}\mid X^{(n)}\right).
    \label{eq:tam_mle}
\end{equation}
Under a fixed-variance isotropic Gaussian observation model with conditional mean $F_{\theta}(X)$, maximizing this likelihood is equivalent to minimizing the mean squared prediction error:
\begin{equation}
    \mathcal L_{\mathrm{sup}}(\theta)=\frac{1}{ND}\sum_{n=1}^{N}
    \left\|Y^{(n)}-F_{\theta}(X^{(n)})\right\|_F^2,
    \label{eq:tam_supervised}
\end{equation}
where $D$ is the number of scalar elements in each target sequence. The predictor $F_{\theta}(X)$ provides the deterministic forecast used at inference.

A pretrained teacher $F_t$ and a student $F_s$ produce forecasts $\widehat Y_t=F_t(X)$ and $\widehat Y_s=F_s(X)$, with intermediate features $Z_t$ and $Z_s$. Distillation augments the student's supervised objective with output and optional feature matching, while keeping the teacher fixed. We further introduce a memory $\mathcal M$ to transfer relationships to previous training samples. Its construction specifies both the predictive representation and the historical references used to supervise it.

\Needspace{6\baselineskip}
\subsection{Task-Aware Teacher Memory}
\label{sec:tam_history}

\paragraph{Predictive representations.}
Different tasks expose predictive structure through different quantities: latent features encode spatial states, forecast differences describe temporal changes, and flow residuals measure departures from persistence. We write the corresponding representation as
\begin{equation}
    u_i^a=\phi_a(X,Z_a,\widehat Y_a)_i,
    \qquad a\in\{t,s\},
    \label{eq:tam_representation}
\end{equation}
where $i$ indexes a region, a spatial location, or a sequence. The student transformation includes a learned adapter when dimensions differ. Relation matching uses normalized vectors $q_i^a=u_i^a/\lVert u_i^a\rVert_2$ and normalized teacher references. Residual regression instead preserves magnitudes to define targets in prediction space.

\paragraph{Reference selection.}
Historical entries are useful only when their comparison is meaningful for the task. We express reference selection as
\begin{equation}
    \mathcal R_i=\operatorname{Select}(\mathcal M;X,u_i^t,\gamma_i),
    \label{eq:tam_selection}
\end{equation}
where $\gamma_i$ denotes metadata such as geographic location or episode identity; each instantiation uses the arguments it requires. Observed motion or foreground occupancy selects video anchors, geographic location restricts weather references, and observation keys retrieve sequences with related input conditions. Eligible entries are sampled or ranked by teacher-feature or observation-key similarity. Appendix~\ref{sec:memory_details} specifies the representations and selection rules for each task.

\subsection{Distillation over Shared Historical References}
\label{sec:tam_relation}

Following cross-image relational distillation~\citep{yang2022cirkd}, the teacher and student define similarity distributions over a shared set of historical teacher vectors. For normalized teacher references $v_j$ and temperature $\tau$, the distributions are
\begin{equation}
    p_{i,j}^{a}
    =\frac{\exp((q_i^{a})^\top v_j/\tau)}
           {\sum_{\ell\in\mathcal R_i}\exp((q_i^{a})^\top v_\ell/\tau)},
    \qquad a\in\{t,s\},\quad j\in\mathcal R_i.
    \label{eq:tam_distribution}
\end{equation}
Using the same references gives corresponding probability components the same meaning. We transfer this relational structure by minimizing
\begin{equation}
    \mathcal L_{\mathrm{rel}}
      =\sum_{i\in\mathcal A}w_i\,
        D_{\mathrm{KL}}\!\left(\operatorname{sg}(p_i^t)\,\Vert\,p_i^s\right),
    \label{eq:tam_relation}
\end{equation}
where $\mathcal A$ is the anchor set, $\operatorname{sg}$ denotes stop-gradient, and the nonnegative weights satisfy $\sum_{i\in\mathcal A}w_i=1$. Video variants average anchors uniformly; geographic-grid variants can use latitude-dependent weights. This objective supervises relative similarity to historical samples, complementing the direct matching of current outputs and features.

\subsection{Observation-Conditioned Residual Prototypes}
\label{sec:tam_prototype}

For traffic flow prediction, we use historical teacher forecasts to construct targets in prediction space. Taking the latest observation $X_T$ as a persistence reference, the residual $\widehat Y_{a,h}-X_T$ represents the predicted change at horizon $h$. We pool and concatenate these residual fields separately over near- and far-horizon groups, yielding vectors $r_a^g$ for $g\in\{\mathrm{near},\mathrm{far}\}$ and $a\in\{t,s\}$. Keeping the groups separate preserves their horizon-specific targets.

Each memory entry stores the teacher residuals together with an observation key $k(X)$. The key is the normalized concatenation of the spatially pooled latest observation $X_T$ and its recent change $X_T-X_{T-1}$. Retrieval therefore depends only on observed inputs. Let $\mathcal R(X)$ contain the top-$k$ historical keys by cosine similarity, and let $\alpha_j(X)$ be the softmax weight of $k(X)^\top k_j/\tau$ over this set. The residual prototype is
\begin{equation}
    \bar r^g(X)=\sum_{j\in\mathcal R(X)}\alpha_j(X)\,r_{t,j}^g,
    \qquad g\in\{\mathrm{near},\mathrm{far}\}.
    \label{eq:tam_prototype}
\end{equation}
Both groups share the retrieved neighbors and weights. The student regresses their respective prototypes through
\begin{equation}
    \mathcal L_{\mathrm{proto}}
      =\frac{1}{2}\sum_{g\in\{\mathrm{near},\mathrm{far}\}}
       \operatorname{MSE}\!\left(r_s^g,\operatorname{sg}(\bar r^g(X))\right).
    \label{eq:tam_prototype_loss}
\end{equation}
Thus, historical forecasts under similar observed conditions provide an additional target for the student's predicted changes. Unlike relation matching, this objective transfers residual values directly. Appendix~\ref{sec:memory_details} gives the explicit encoding and retrieval definitions.

\subsection{Joint Optimization and Memory Updates}
\label{sec:tam_optimization}

We combine historical supervision with ground-truth prediction, output distillation, and optional feature alignment:
\begin{equation}
    \mathcal L=\mathcal L_{\mathrm{sup}}
      +\lambda_{\mathrm{out}}\mathcal L_{\mathrm{out}}
      +\lambda_{\mathrm{feat}}\mathcal L_{\mathrm{feat}}
      +\lambda_{\mathrm{mem}}\mathcal L_{\mathrm{mem}}.
    \label{eq:tam_total}
\end{equation}
The supervised term is $\mathcal L_{\mathrm{sup}}=\operatorname{MSE}(\widehat Y_s,Y)$, and output distillation uses $\mathcal L_{\mathrm{out}}=\operatorname{MSE}(\widehat Y_s,\operatorname{sg}(\widehat Y_t))$. MSE averages over tensor elements. The optional feature term aligns dimension-adapted intermediate features, while $\mathcal L_{\mathrm{mem}}$ is the task's relation or prototype loss. Training jointly optimizes the student and enabled adapters, with losses aggregated over the batch.

Each iteration retrieves references and evaluates the memory loss before inserting the current detached teacher entries. The bounded memory retains recent training history under the task's location or episode constraints. Before sufficient history is available, optimization uses only prediction and conventional distillation losses. Memory order follows training iterations and does not imply chronological ordering of shuffled sequences.

At inference, $\widehat Y_s=F_s(X)$ requires only the student. The teacher, memory, and auxiliary adapters are used exclusively during training and add no student inference cost.

\begingroup
\renewcommand{\floatpagefraction}{0.85}
\setlength{\textfloatsep}{12pt plus 2pt minus 2pt}
\setlength{\intextsep}{10pt plus 2pt minus 2pt}
\setlength{\floatsep}{10pt plus 2pt minus 2pt}
\section{Experiments}
\label{sec:experiments}

\subsection{Experimental Setup}
\label{sec:experimental_setup}

\paragraph{Datasets and protocols.}
We evaluate \method{} on video prediction, weather forecasting, and traffic flow prediction. Video benchmarks include Moving MNIST~\citep{srivastava2015video}, Moving FMNIST constructed from Fashion-MNIST~\citep{xiao2017fashion}, KTH~\citep{schuldt2004actions}, Human3.6M~\citep{ionescu2014human36m}, HMDB51~\citep{kuehne2011hmdb}, BAIR~\citep{ebert2017visualplanning}, and KittiCaltech, which pairs KITTI~\citep{geiger2013kitti} training data with Caltech Pedestrian~\citep{dollar2009pedestrian} test videos. Weather and traffic tasks use WeatherBench two-meter temperature~\citep{rasp2020weatherbench} and TaxiBJ inflow/outflow fields~\citep{zhang2017stresnet}. KTH and Human3.6M use subject-disjoint splits; HMDB51 follows official split~1; and BAIR uses our episode-balanced window-sampling protocol. WeatherBench uses chronological splits. TaxiBJ reserves the final portion of the official training set for validation and removes overlapping boundary windows. Appendix~\ref{sec:experiment_protocols} specifies our prediction, preprocessing, and evaluation protocols.

\paragraph{Models and comparisons.}
Teachers use the OpenSTL forecasting implementation~\citep{tan2023openstl}, with TAU~\citep{tan2023tau}, IncepU~\citep{gao2022simvp}, gSTA~\citep{tan2025simvpv2}, or UniFormer~\citep{li2022uniformer} translators. Students are forecasting adaptations of U-Net~\citep{ronneberger2015unet}, with Base and Tiny variants, and a ResNet-18-style encoder~\citep{he2016resnet} with a fully convolutional decoder (ResNet-FCN). We compare supervised-only training, output distillation, output plus standardized feature-MSE distillation, and output plus activation-boundary distillation~\citep{heo2019activation}. \method{} adds historical supervision to the corresponding distillation objective. Memory comparisons retain the same teacher, student, and underlying distillation configuration. Teachers remain frozen during training; inference uses only \mbox{the student network}.

\paragraph{Implementation.}
All configurations use Adam~\citep{kingma2015adam}. Video and traffic tasks use a one-cycle schedule~\citep{smith2018onecycle} with a peak learning rate of $10^{-3}$; WeatherBench uses cosine decay~\citep{loshchilov2017sgdr} with an initial rate of $5\times10^{-3}$. Training budgets are 200 epochs for Moving MNIST, Moving FMNIST, and BAIR, 50 for WeatherBench, and 100 for the remaining tasks. Batch size is 16, except for KittiCaltech, which uses 8. The output-distillation weight is 1; feature and memory weights depend on the task. Standard relational configurations maintain 4,096 teacher entries, enqueue up to 40 per iteration, and sample 1,024 references. BAIR, WeatherBench, and TaxiBJ instead use episode-aware retrieval, location-specific history, and observation-conditioned residual prototypes, respectively. Appendix~\ref{sec:training_details} provides the configurations and hyperparameters.

\paragraph{Evaluation.}
Video metrics are MSE, MAE, PSNR, SSIM~\citep{wang2004image}, and LPIPS~\citep{zhang2018unreasonable}. TaxiBJ uses MSE, MAE, and RMSE; WeatherBench uses normalized errors and both ordinary and latitude-weighted RMSE in Kelvin. Tables specify the error reduction and scaling conventions. Multi-seed experiments use seeds 42--45 and report means, standard deviations, and the actual number of runs; relative changes use matched-seed comparisons. Tasks with an independent validation set select checkpoints by validation MSE. The current Moving MNIST, Moving FMNIST, KTH, and Human3.6M implementations share validation and test data; their evaluation-set checkpoint selection is reported separately, alongside available final-epoch results. A dagger in Table~\ref{tab:video_main} denotes this shared-split protocol.

\subsection{Comparison with Distillation Baselines}
\label{sec:video_results}

Table~\ref{tab:video_main} compares supervised training, conventional knowledge distillation, and the addition of \method{} across teacher--student configurations. With the teacher, student architecture, and base distillation objective fixed, \method{} improves mean SSIM on all six datasets and reduces mean MSE on five. On KittiCaltech, for example, it reduces MSE from 6.983 to 6.853 (in the table's $10^3$ scaling), raises SSIM from 0.6873 to 0.6917, and lowers LPIPS from 0.4199 to 0.4128. These gains indicate that historical teacher information provides useful supervision for cross-domain prediction.

\begin{table}[!t]
\begingroup
\normalsize
\fontfamily{EBGaramond-LF}\selectfont
\definecolor{TAMTableInk}{HTML}{245F69}
\definecolor{TAMTableBand}{HTML}{EDF3F3}
\definecolor{TAMTableOurs}{HTML}{F3EFFB}
\captionsetup{font={normalsize,tamgaramond},labelfont=bf,skip=6pt}
\caption{Video prediction under different teacher--student configurations. Students report mean $\pm$ population standard deviation over four runs; teachers use seed~42. Elementwise MSE is scaled by $10^3$. Bold denotes the best student mean in each configuration.}
\label{tab:video_main}
\centering
\setlength{\tabcolsep}{4pt}
\renewcommand{\arraystretch}{1.02}
\arrayrulecolor{TAMTableInk}
\begin{tabularx}{\linewidth}{@{}Xrrrr@{}}
\toprule[1pt]
\textbf{Method} & \textbf{Params (M)} & \textbf{MSE} $\downarrow$ & \textbf{SSIM} $\uparrow$ & \textbf{LPIPS} $\downarrow$ \\
\midrule
\rowcolor{TAMTableBand}[0pt][0pt]
\multicolumn{5}{@{}p{\linewidth}@{}}{\rule{0pt}{12pt}\hspace{4pt}\textcolor{TAMTableInk}{\textbf{KittiCaltech}}\hfill UniFormer $\rightarrow$ ResNet-FCN\hspace{4pt}} \\
Teacher & 11.787 & $2.124$ & $0.9099$ & $0.0661$ \\
Student & 12.811 & $7.187\,\pm\,0.251$ & $0.6813\,\pm\,0.0066$ & $0.4332\,\pm\,0.0093$ \\
Output KD & 12.811 & $6.983\,\pm\,0.087$ & $0.6873\,\pm\,0.0034$ & $0.4199\,\pm\,0.0138$ \\
\rowcolor{TAMTableOurs}
\textbf{KD + TAM} & 12.811 & $\mathbf{6.853}\,\pm\,0.112$ & $\mathbf{0.6917}\,\pm\,0.0019$ & $\mathbf{0.4128}\,\pm\,0.0078$ \\
\addlinespace[2pt]
\rowcolor{TAMTableBand}[0pt][0pt]
\multicolumn{5}{@{}p{\linewidth}@{}}{\rule{0pt}{12pt}\hspace{4pt}\textcolor{TAMTableInk}{\textbf{HMDB51}}\hfill TAU $\rightarrow$ U-Net-Base\hspace{4pt}} \\
Teacher & 10.607 & $3.285$ & $0.8831$ & $0.0597$ \\
Student & 7.706 & $4.331\,\pm\,0.022$ & $0.8371\,\pm\,0.0060$ & $0.0638\,\pm\,0.0017$ \\
Feature-MSE KD & 7.706 & $\mathbf{3.829}\,\pm\,0.025$ & $0.8472\,\pm\,0.0031$ & $\mathbf{0.0625}\,\pm\,0.0011$ \\
\rowcolor{TAMTableOurs}
\textbf{KD + TAM} & 7.706 & $3.843\,\pm\,0.036$ & $\mathbf{0.8525}\,\pm\,0.0018$ & $0.0635\,\pm\,0.0008$ \\
\addlinespace[2pt]
\rowcolor{TAMTableBand}[0pt][0pt]
\multicolumn{5}{@{}p{\linewidth}@{}}{\rule{0pt}{12pt}\hspace{4pt}\textcolor{TAMTableInk}{\textbf{BAIR}}\hfill TAU $\rightarrow$ Tiny U-Net\hspace{4pt}} \\
Teacher & 10.607 & $23.294$ & $0.6706$ & $0.2317$ \\
Student & 0.119 & $\mathbf{24.556}\,\pm\,0.332$ & $0.6419\,\pm\,0.0046$ & $0.2454\,\pm\,0.0020$ \\
Output KD & 0.119 & $24.773\,\pm\,0.374$ & $0.6439\,\pm\,0.0045$ & $\mathbf{0.2406}\,\pm\,0.0005$ \\
\rowcolor{TAMTableOurs}
\textbf{KD + TAM} & 0.119 & $24.621\,\pm\,0.225$ & $\mathbf{0.6455}\,\pm\,0.0029$ & $0.2423\,\pm\,0.0025$ \\
\addlinespace[2pt]
\rowcolor{TAMTableBand}[0pt][0pt]
\multicolumn{5}{@{}p{\linewidth}@{}}{\rule{0pt}{12pt}\hspace{4pt}\textcolor{TAMTableInk}{\textbf{KTH$^{\dagger}$}}\hfill IncepU $\rightarrow$ U-Net-Base\hspace{4pt}} \\
Teacher & 12.152 & $2.544$ & $0.8146$ & $0.2755$ \\
Student & 7.705 & $2.673\,\pm\,0.013$ & $0.8120\,\pm\,0.0011$ & $\mathbf{0.2493}\,\pm\,0.0053$ \\
Feature-MSE KD & 7.705 & $2.623\,\pm\,0.051$ & $0.8123\,\pm\,0.0024$ & $0.2693\,\pm\,0.0042$ \\
\rowcolor{TAMTableOurs}
\textbf{KD + TAM} & 7.705 & $\mathbf{2.605}\,\pm\,0.024$ & $\mathbf{0.8132}\,\pm\,0.0006$ & $0.2652\,\pm\,0.0011$ \\
\addlinespace[2pt]
\rowcolor{TAMTableBand}[0pt][0pt]
\multicolumn{5}{@{}p{\linewidth}@{}}{\rule{0pt}{12pt}\hspace{4pt}\textcolor{TAMTableInk}{\textbf{Moving MNIST$^{\dagger}$}}\hfill TAU $\rightarrow$ ResNet-FCN\hspace{4pt}} \\
Teacher & 44.656 & $6.092$ & $0.9340$ & $0.0557$ \\
Student & 12.748 & $15.033\,\pm\,0.087$ & $0.7994\,\pm\,0.0113$ & $0.2382\,\pm\,0.0044$ \\
Output KD & 12.748 & $14.845\,\pm\,0.052$ & $0.7990\,\pm\,0.0085$ & $0.2347\,\pm\,0.0035$ \\
\rowcolor{TAMTableOurs}
\textbf{KD + TAM} & 12.748 & $\mathbf{14.811}\,\pm\,0.041$ & $\mathbf{0.8026}\,\pm\,0.0066$ & $\mathbf{0.2341}\,\pm\,0.0022$ \\
\addlinespace[2pt]
\rowcolor{TAMTableBand}[0pt][0pt]
\multicolumn{5}{@{}p{\linewidth}@{}}{\rule{0pt}{12pt}\hspace{4pt}\textcolor{TAMTableInk}{\textbf{Moving FMNIST$^{\dagger}$}}\hfill TAU $\rightarrow$ ResNet-FCN\hspace{4pt}} \\
Teacher & 44.656 & $5.999$ & $0.8767$ & $0.1413$ \\
Student & 12.748 & $11.981\,\pm\,0.011$ & $0.7455\,\pm\,0.0034$ & $0.4124\,\pm\,0.0024$ \\
Output KD & 12.748 & $11.844\,\pm\,0.013$ & $0.7473\,\pm\,0.0022$ & $0.4074\,\pm\,0.0028$ \\
\rowcolor{TAMTableOurs}
\textbf{KD + TAM} & 12.748 & $\mathbf{11.835}\,\pm\,0.017$ & $\mathbf{0.7483}\,\pm\,0.0036$ & $\mathbf{0.4059}\,\pm\,0.0034$ \\
\addlinespace[2pt]
\bottomrule[1pt]
\end{tabularx}
\arrayrulecolor{black}

\endgroup
\end{table}

\method{} improves all three metrics over the corresponding KD baseline on KittiCaltech, KTH, Moving MNIST, and Moving FMNIST. The gains nevertheless vary across metrics: HMDB51 and BAIR achieve higher SSIM without a corresponding improvement in LPIPS.

These results support teacher memory as a complement to existing distillation objectives. Student parameter counts remain unchanged, and inference requires neither the teacher nor memory retrieval.

\subsection{Weather and Traffic Forecasting}
\label{sec:cross_task_results}

Table~\ref{tab:cross_task} extends the evaluation to weather and traffic forecasting. On WeatherBench, \method{} lowers mean MSE, MAE, and RMSE relative to Feature-MSE KD with both teachers. With gSTA, paired MSE decreases by $1.93\%$ on average and temperature RMSE falls from 1.1191\,K to 1.1081\,K; TAU yields smaller gains. Geographically organized teacher history thus provides additional supervision, but its benefit depends on the teacher and does not consistently reduce variance.

\begin{table}[!htbp]
\begingroup
\normalsize
\fontfamily{EBGaramond-LF}\selectfont
\definecolor{TAMTableInk}{HTML}{245F69}
\definecolor{TAMTableBand}{HTML}{EDF3F3}
\definecolor{TAMTableOurs}{HTML}{F3EFFB}
\captionsetup{font={normalsize,tamgaramond},labelfont=bf,skip=6pt}
\caption{Weather and traffic forecasting (mean $\pm$ population SD over four runs). TAM augments Feature-MSE KD; Feature-AB KD is an independent baseline. Bold marks the best student mean per group. MSE/MAE are normalized; RMSE units are specified per group.}
\label{tab:cross_task}
\centering
\setlength{\tabcolsep}{4pt}
\renewcommand{\arraystretch}{1.04}
\arrayrulecolor{TAMTableInk}
\begin{tabularx}{\linewidth}{Xrrr}
\toprule[1pt]
\textbf{Method} & \textbf{MSE} $\times10^3\downarrow$ & \textbf{MAE} $\times10^3\downarrow$ & \textbf{RMSE} $\downarrow$ \\
\midrule
\rowcolor{TAMTableBand}[0pt][0pt]
\multicolumn{4}{@{}p{\linewidth}@{}}{\rule{0pt}{12pt}\hspace{4pt}\textcolor{TAMTableInk}{\textbf{WeatherBench $\cdot$ TAU $\rightarrow$ U-Net-Base}}\hfill RMSE (K)\hspace{4pt}} \\
Student & $2.833\,\pm\,0.012$ & $33.798\,\pm\,0.211$ & $1.1272\,\pm\,0.0024$ \\
Feature-MSE KD & $2.681\,\pm\,0.025$ & $32.419\,\pm\,0.288$ & $1.0965\,\pm\,0.0051$ \\
\rowcolor{TAMTableOurs}
\textbf{MSE KD + TAM} & $\mathbf{2.679}\,\pm\,0.016$ & $\mathbf{32.326}\,\pm\,0.161$ & $\mathbf{1.0962}\,\pm\,0.0034$ \\
\addlinespace[3pt]
\rowcolor{TAMTableBand}[0pt][0pt]
\multicolumn{4}{@{}p{\linewidth}@{}}{\rule{0pt}{12pt}\hspace{4pt}\textcolor{TAMTableInk}{\textbf{WeatherBench $\cdot$ gSTA $\rightarrow$ U-Net-Base}}\hfill RMSE (K)\hspace{4pt}} \\
Student & $2.833\,\pm\,0.012$ & $33.798\,\pm\,0.211$ & $1.1272\,\pm\,0.0024$ \\
Feature-MSE KD & $2.793\,\pm\,0.098$ & $33.530\,\pm\,0.842$ & $1.1191\,\pm\,0.0197$ \\
\rowcolor{TAMTableOurs}
\textbf{MSE KD + TAM} & $\mathbf{2.740}\,\pm\,0.166$ & $\mathbf{32.961}\,\pm\,1.432$ & $\mathbf{1.1081}\,\pm\,0.0330$ \\
\addlinespace[3pt]
\rowcolor{TAMTableBand}[0pt][0pt]
\multicolumn{4}{@{}p{\linewidth}@{}}{\rule{0pt}{12pt}\hspace{4pt}\textcolor{TAMTableInk}{\textbf{TaxiBJ $\cdot$ TAU $\rightarrow$ U-Net-Base}}\hfill RMSE $\times10^3$\hspace{4pt}} \\
Student & $0.2123\,\pm\,0.0034$ & $8.146\,\pm\,0.137$ & $14.571\,\pm\,0.118$ \\
Feature-MSE KD & $0.2002\,\pm\,0.0034$ & $8.052\,\pm\,0.145$ & $14.148\,\pm\,0.121$ \\
\rowcolor{TAMTableOurs}
\textbf{MSE KD + TAM} & $0.1981\,\pm\,0.0054$ & $7.949\,\pm\,0.061$ & $14.073\,\pm\,0.193$ \\
Feature-AB KD & $\mathbf{0.1881}\,\pm\,0.0039$ & $\mathbf{7.830}\,\pm\,0.045$ & $\mathbf{13.716}\,\pm\,0.141$ \\
\bottomrule[1pt]
\end{tabularx}
\arrayrulecolor{black}
\endgroup
\end{table}

On TaxiBJ, observation-conditioned residual memory lowers MSE by $1.01\%$ on average over paired Feature-MSE KD runs, while also reducing MAE and RMSE. These gains support historical residual supervision for traffic prediction. Feature-AB KD remains stronger, however, so \method{} complements the matched baseline without outperforming every distillation objective. Across video, weather, and traffic prediction, mean gains vary with the underlying distillation configuration.

\subsection{Prediction Horizon Analysis}
\label{sec:horizon_analysis}

Figure~\ref{fig:horizon_gains} reports horizon-dependent gains over matched KD on BAIR and TaxiBJ.

\begin{figure}[!htbp]
    \centering
    \includegraphics[width=\linewidth]{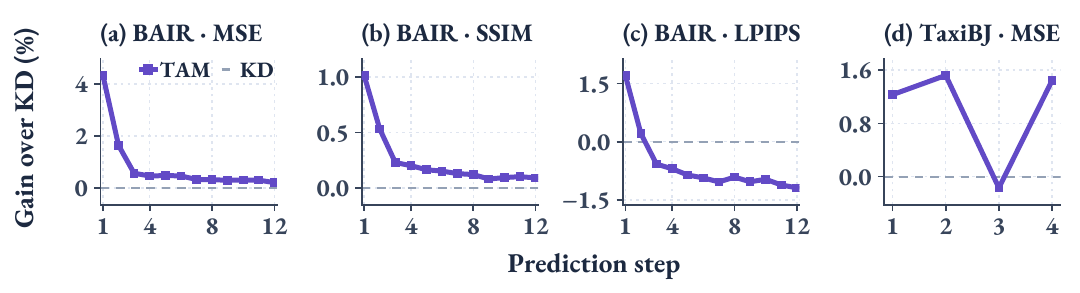}
    \caption{Prediction-horizon gains over matched KD baselines, averaged over four paired runs. Positive values indicate lower MSE/LPIPS or higher SSIM; the dashed line denotes the KD baseline.}
    \label{fig:horizon_gains}
\end{figure}

On BAIR, \method{} improves MSE and SSIM at all 12 prediction steps, but the MSE reduction falls from $4.33\%$ at the first step to $0.20\%$ at the last, indicating stronger near-term benefits. LPIPS improves only at the first two steps, so pixel and structural gains do not imply uniformly better perceptual quality. TaxiBJ exhibits MSE reductions of $1.23\%$, $1.52\%$, $-0.17\%$, and $1.44\%$ across its four steps. Thus, gains vary across horizons and are not monotonic.

\subsection{Spatial Frequency Analysis}
\label{sec:spectral_analysis}

Figure~\ref{fig:spectral_analysis} compares U-Net-Base power spectra and frequency-resolved errors for supervised training (S0) and \method{}. Results cover all forecast frames from 256 fixed, evenly spaced test windows per dataset and four paired runs. Appendix~\ref{sec:spectral_protocol} defines the transform and frequency bands.

\begin{figure}[!htbp]
    \centering
    \includegraphics[width=\linewidth]{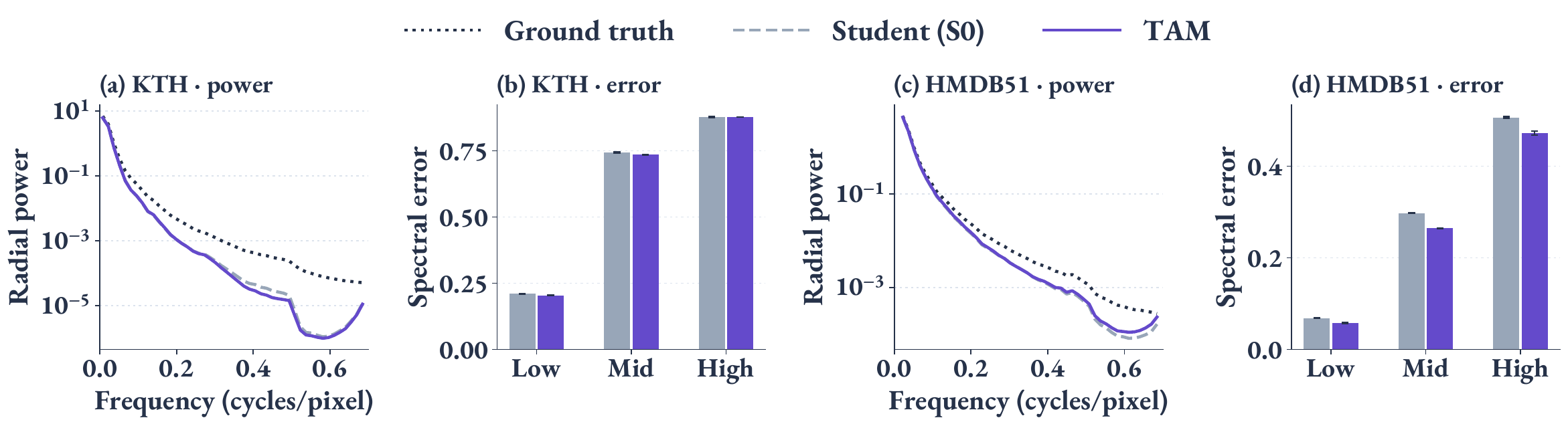}
    \caption{Spatial frequency analysis on KTH and HMDB51. Lower normalized spectral error is better; bars and error bars show means and standard deviations over four training runs.}
    \label{fig:spectral_analysis}
\end{figure}

On HMDB51, \method{} reduces spectral error relative to S0 by $15.44\%$, $11.14\%$, and $6.73\%$ in the low-, mid-, and high-frequency bands, respectively. KTH exhibits smaller reductions of $3.00\%$ and $1.15\%$ in the first two bands, while high-frequency error remains essentially unchanged (a $0.07\%$ increase). Both students retain less high-frequency power than GT. The analysis therefore indicates dataset-dependent improvements in spatial reconstruction, with larger relative gains at low frequencies, rather than uniform recovery of fine detail.

\subsection{Motion-Conditioned Prediction Analysis}
\label{sec:motion_analysis}

Figure~\ref{fig:motion_analysis} compares \method{} and S0 across motion conditions on 1,024 fixed windows per dataset and four paired runs. Absolute differences between the final two input frames define shared spatial masks; their spatial means determine motion groups (Appendix~\ref{sec:motion_protocol}).

\begin{figure}[!htbp]
    \centering
    \includegraphics[width=\linewidth]{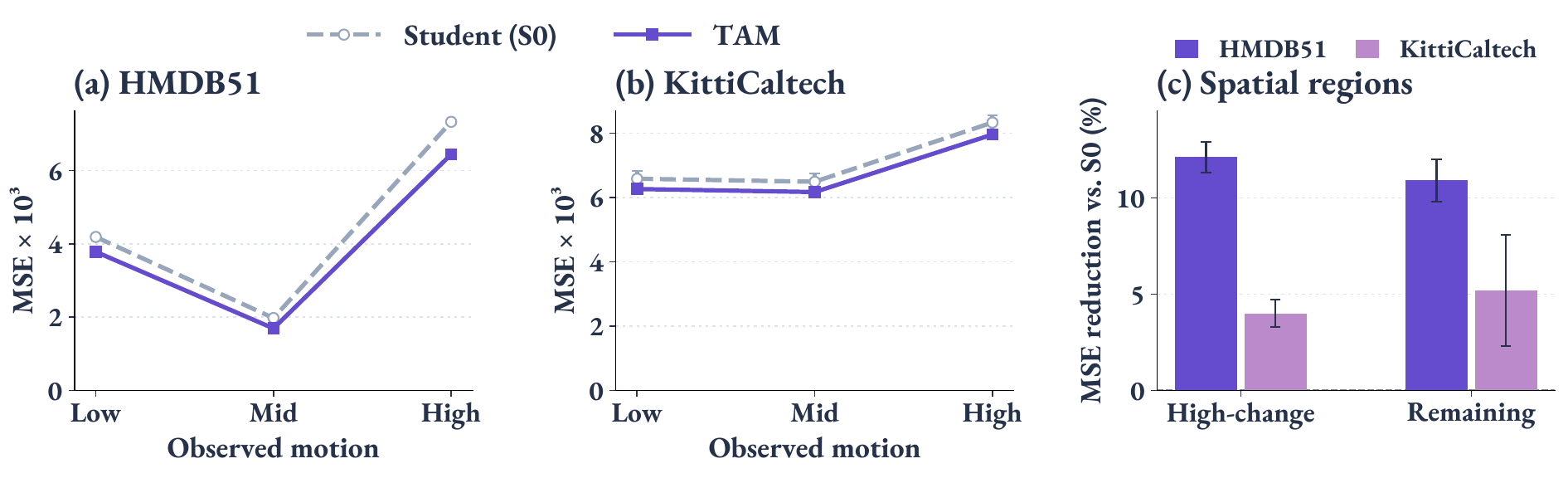}
    \caption{Prediction errors across observed-motion groups (left, middle) and relative MSE reductions in input-defined spatial regions (right). Error bars show standard deviations over four paired runs.}
    \label{fig:motion_analysis}
\end{figure}

\method{} reduces MSE by $9.68\%$, $14.31\%$, and $12.12\%$ in the low-, middle-, and high-motion HMDB51 groups, and by $4.78\%$, $4.85\%$, and $4.45\%$ on KittiCaltech. High-change regions improve by $12.12\%$ and $4.00\%$, respectively, although KittiCaltech benefits more in the remaining region ($5.19\%$). Gains thus extend across motion conditions without increasing monotonically with motion strength. These comparisons assess the full framework; Table~\ref{tab:video_main} tests memory against matched KD.

\endgroup

\section{Conclusion}
\label{sec:conclusion}

We present \method{}, a task-aware memory distillation framework that turns a frozen teacher's historical predictions and representations into cross-sample supervision. Task-specific representations and reference selection support relation alignment over shared history or observation-conditioned residual prototype regression, complementing per-sample distillation. Experiments on video, weather, and traffic prediction demonstrate improvements across multiple teacher--student configurations without additional student inference cost. Prediction-horizon, frequency, and motion-conditioned analyses show that gains vary across forecast steps, spatial frequencies, and motion conditions; improvements in pixel and structural metrics do not necessarily yield consistent perceptual gains. These findings support the organization of historical references as an important design choice for predictive distillation. The current framework still relies on task-specific representations and retrieval rules, and the available comparisons do not fully isolate the contributions of individual design choices. Future work will investigate more unified reference selection and controlled ablations to clarify the roles of different memory designs.

\clearpage
\section*{AI Use Statement}
We use ChatGPT and Codex to assist with literature search and synthesis, refinement of the research framework and its mathematical presentation, manuscript drafting and translation between Chinese and English, implementation and debugging of inference and analysis scripts, discussion of experimental and post-hoc analysis designs, interpretation of results, and figure creation and typesetting. The authors review and verify the AI-assisted work and take responsibility for the final text, code, experimental results, and scientific conclusions.

\section*{Reproducibility Statement}
Section~\ref{sec:method} and Appendix~\ref{sec:memory_details} describe the method and task-specific memory designs. Section~\ref{sec:experimental_setup} and Appendices~\ref{sec:experiment_protocols}--\ref{sec:training_details} document preprocessing, data splits, checkpoint selection, model configurations, training hyperparameters, and evaluation rules. Appendices~\ref{sec:spectral_protocol}--\ref{sec:motion_protocol} specify the sampling and computation procedures for frequency, motion-conditioned, and temporal analyses. We report the number of repeated runs, statistical conventions, and paired comparisons, and explicitly identify settings that share validation and test data. Code and experimental configurations will be released.

\begingroup
\urlstyle{same}
\providecommand{\doi}[1]{}
\renewcommand{\doi}[1]{doi:\nobreakspace\mbox{#1}}
\bibliography{iclr2026_conference}
\endgroup
\bibliographystyle{tam_layout}
\clearpage
\appendix

\numberwithin{equation}{section}
\section{Task-Specific Memory Instantiations}
\label{sec:memory_details}
\renewcommand{\thetable}{A\arabic{table}}
\setcounter{table}{0}

The implementation instantiates the memory interface with the representations and reference policies summarized in Table~\ref{tab:tam_memory}. These settings specialize the training objective to each task; the prediction network does not access the memory at inference.

\begin{table}[!htbp]
    \centering
    \caption{Memory instantiations. ``Region $r$'' denotes non-overlapping $r\times r$ average pooling. Representations and anchor selection are listed separately from reference organization.}
    \label{tab:tam_memory}
    \small
    \setlength{\tabcolsep}{4pt}
    \renewcommand{\arraystretch}{1.12}
    \begin{tabularx}{\linewidth}{@{}>{\raggedright\arraybackslash}p{.18\linewidth}>{\raggedright\arraybackslash}X>{\raggedright\arraybackslash}X@{}}
    \toprule
    Task & Representation / current anchors & Historical references \\
    \midrule
    H36M & Region-2 features; observed-motion mask & 4,096-token FIFO; sample 1,024 \\
    HMDB51 & Features at high-motion locations & 4,096-token FIFO; sample 1,024 \\
    BAIR & One motion-pooled feature token \mbox{for each sequence} & Episode keys; retrieve 64 \mbox{from other episodes} \\
    KittiCaltech & Region-2 features; all regions are anchors & Sample 1,024; enqueue \mbox{high- and ordinary-motion tokens} \\
    Moving MNIST / FMNIST & Region-2 features; recent-occupancy mask & Sample 1,024; retain top-128 \mbox{teacher references per anchor} \\
    KTH & Regional forecast differences with coordinates; motion anchors & 4,096-token FIFO; sample 1,024 \\
    WeatherBench & Features at each geographic grid location & 64 historical windows per location \\
    TaxiBJ & Pooled residuals from the latest flow; \mbox{separate near/far horizons} & 2,048 observation-keyed entries; retrieve 16 \\
    \bottomrule
    \end{tabularx}
\end{table}

Observed motion or occupancy selects video anchors, while the representation and reference policy determine the transferred relation. BAIR uses one motion-pooled token per sequence, retains at most one historical entry per episode, and excludes the current episode from retrieval. KTH augments regional forecast differences with spatial coordinates. Weather memory compares features at corresponding geographic locations and weights the loss by normalized cosine latitude. Appendix~\ref{sec:training_details} lists pooling sizes, memory capacities, warm-up requirements, and other hyperparameters.

\paragraph{Residual prototype definitions.}
Let $\mathcal P$ apply spatial average pooling to each field and then vectorize the result, preserving the horizon and channel order. With chronological observation notation, the key and residual encodings in Section~\ref{sec:tam_prototype} are
\begin{align}
    k(X)&=\operatorname{norm}_2\!\left(\mathcal P([X_T,X_T-X_{T-1}])\right),
    \label{eq:tam_key}\\
    r_a^g&=\mathcal P\!\left([\widehat Y_{a,h}-X_T]_{h\in\mathcal H_g}\right),
    \qquad a\in\{t,s\},\quad g\in\{\mathrm{near},\mathrm{far}\}.
    \label{eq:tam_residual}
\end{align}
Here brackets denote concatenation. The TaxiBJ loader stores the latest observation first; the notation above orders observations chronologically. We use $\mathcal H_{\mathrm{near}}=\{1,2\}$ and $\mathcal H_{\mathrm{far}}=\{3,4\}$ for TaxiBJ. Pooling and concatenation retain individual horizons within each group. The normalized observation keys determine the reference weights:
\begin{equation}
    \alpha_j(X)=\frac{\exp(k(X)^\top k_j/\tau)}
    {\sum_{\ell\in\mathcal R(X)}\exp(k(X)^\top k_\ell/\tau)},
    \qquad j\in\mathcal R(X).
    \label{eq:tam_prototype_weights}
\end{equation}
The same weights aggregate the near- and far-horizon residual banks; both targets are detached during student optimization.

The global-residual control in Figure~\ref{fig:tam_pilot} retrieves with the student's normalized predicted residual and combines all future horizons into one vector. Comparing it with observation-conditioned near/far prototypes changes both the retrieval key and horizon organization, without isolating \mbox{their individual effects}.

\clearpage
\section{Data and Evaluation Protocols}
\label{sec:experiment_protocols}

Table~\ref{tab:dataset_protocols} summarizes the prediction tasks. A time step denotes a sampled video frame, an hourly temperature field, or a traffic observation, depending on the dataset. Image values are scaled to $[0,1]$; temperature fields use training-set standardization.

\begin{table}[!htbp]
    \centering
    \caption{Prediction protocols. Channels refer to the predicted field; Human3.6M retains the BGR channel order of its preprocessing pipeline.}
    \label{tab:dataset_protocols}
    \small
    \setlength{\tabcolsep}{7pt}
    \renewcommand{\arraystretch}{1.12}
    \begin{tabularx}{\linewidth}{@{}Xccc@{}}
    \toprule
    Dataset & Input $\to$ output & Resolution & Channels \\
    \midrule
    Moving MNIST / FMNIST & $10\to10$ & $64\times64$ & 1 \\
    KTH & $10\to20$ & $128\times128$ & 1 \\
    Human3.6M & $4\to4$ & $256\times256$ & 3 \\
    HMDB51 & $4\to4$ & $64\times64$ & 3 \\
    BAIR & $4\to12$ & $64\times64$ & 3 \\
    KittiCaltech & $10\to1$ & $128\times160$ & 3 \\
    WeatherBench & $12\to12$ & $32\times64$ & 1 \\
    TaxiBJ & $4\to4$ & $32\times32$ & 2 \\
    \bottomrule
    \end{tabularx}
\end{table}

\paragraph{Video data.}
Moving MNIST and Moving FMNIST generate 10,000 training sequences online per epoch, with two moving digit or clothing sprites, and evaluate on fixed sets of 10,000 sequences. KTH uses persons 01--16 for training and 17--25 for testing. Human3.6M uses S1/S5/S6/S7/S8 for training and S9/S11 for testing, sampling every fifth source frame. HMDB51 uses the unstabilized videos of official split~1, with a deterministic, class-stratified 10\% validation holdout from the official training videos; action labels are not prediction targets.

BAIR uses the MP4-based episode-balanced protocol: each epoch samples one window per training episode, and a 15-epoch cycle covers all valid starts. Validation and testing enumerate all windows; the test set contains 3,840 windows from 256 episodes. These results use this sampling and preprocessing protocol rather than the separate full-window TFRecord setup. KittiCaltech uses disjoint KITTI recordings for training and validation, and Caltech Pedestrian test recordings for cross-domain evaluation. Caltech preprocessing samples every third decoded frame, resizes to height 128, and center-crops to width 160.

\paragraph{Weather and traffic data.}
WeatherBench uses two-meter temperature on the $5.625^\circ$ grid. Training, validation, and testing cover 2010--2015, 2016, and 2017--2018, respectively. A scalar mean and standard deviation computed on the training period normalize all splits; windows do not cross temporal gaps. TaxiBJ maps the OpenSTL input range $[-1,1]$ to $[0,1]$. The first 90\% of official training windows form the training partition and the final 10\% form validation, with a seven-window boundary purge. The official test set provides final evaluation.

\paragraph{Checkpoint selection and metric conventions.}
HMDB51, BAIR, KittiCaltech, WeatherBench, and TaxiBJ use independent validation data for MSE-based checkpoint selection. Moving MNIST, Moving FMNIST, KTH, and Human3.6M reuse the evaluation data for validation, so their Best results involve evaluation-set selection; available fixed final-epoch results provide a separate checkpoint criterion. Video PSNR and SSIM use a data range of one. Elementwise mean MSE/MAE and OpenSTL spatial/channel-sum errors are distinguished explicitly; the latter multiply the former by $CHW$ for a fixed field shape. Weather RMSE is also computed after denormalization, and latitude weighting uses normalized cosine latitude. Reported multi-run deviations use the population convention. Relative improvements average per-seed relative changes, rather than taking the ratio of aggregate means; each comparison identifies its metric and reference configuration.

\clearpage
\section{Training Configurations and Hyperparameters}
\label{sec:training_details}

Table~\ref{tab:training_configs} lists the teacher--student configurations underlying the task-specific memory experiments. ``Base'' and ``Tiny'' denote U-Net-Base and Tiny U-Net; ``FCN'' denotes ResNet-FCN. Teachers are trained separately and remain in evaluation mode during student optimization. Output distillation has weight $\lambda_{\mathrm{out}}=1$. A zero feature weight denotes output-only distillation; nonzero entries below use standardized feature MSE. Activation-boundary baselines replace this feature objective while retaining the corresponding task configuration.

\begin{table}[!htbp]
    \centering
    \caption{Training budgets and loss weights for the memory configurations. $E$: epochs; $B$: batch size. Budgets describe configured training lengths, not completion of every archived run.}
    \label{tab:training_configs}
    \small
    \setlength{\tabcolsep}{4pt}
    \renewcommand{\arraystretch}{1.15}
    \begin{tabularx}{\linewidth}{@{}Xllrrrr@{}}
    \toprule
    Task & Teacher & Student & $E$ & $B$ & $\lambda_{\mathrm{feat}}$ & $\lambda_{\mathrm{mem}}$ \\
    \midrule
    Moving MNIST & TAU & FCN & 200 & 16 & 0 & 0.001 \\
    Moving FMNIST & TAU & FCN & 200 & 16 & 0 & 0.001 \\
    KTH & IncepU & Base/Tiny & 100 & 16 & 0.03 & 0.001 \\
    Human3.6M & TAU & Base & 100 & 16 & 0 & 0.001 \\
    HMDB51 & TAU & Base & 100 & 16 & 0.1 & 0.001 \\
    BAIR & TAU & Tiny & 200 & 16 & 0 & 0.001 \\
    KittiCaltech & UniFormer & FCN & 100 & 8 & 0 & 0.003 \\
    WeatherBench & TAU/gSTA & Base & 50 & 16 & 0.1 & 0.001 \\
    WeatherBench & TAU & Tiny & 50 & 16 & 0.1 & 0.01 \\
    TaxiBJ & TAU & Base & 100 & 16 & 0.1 & 0.1 \\
    \bottomrule
    \end{tabularx}
\end{table}

\paragraph{Optimization and representations.}
Adam and OneCycleLR use a peak rate of $10^{-3}$ for video and traffic tasks. WeatherBench instead uses Adam with an initial rate of $5\times10^{-3}$ and cosine decay. U-Net-Base uses four scales with base width 64; Tiny U-Net uses three scales with base width 16. ResNet-FCN uses a ResNet-18-style encoder and a progressive upsampling decoder. KittiCaltech uses the UniFormer-hs64 teacher. Feature adapters map student channels to the teacher dimension before per-sample, per-channel spatial standardization and MSE. Feature and memory adapters are distinct unless explicitly shared by the configuration. Activation-boundary losses average over tensor elements; output-only baselines omit feature alignment. Human3.6M uses FP32; the archived BAIR memory and TaxiBJ memory configurations specify BF16. The historical BAIR output-only configuration does not record precision, so a shared precision setting is not assumed.

\paragraph{Reference construction.}
The standard relational FIFO holds 4,096 entries, enqueues up to 40 per iteration, samples 1,024 references, and starts contributing after the bank fills. Similarity temperature is 0.1 and the auxiliary KD temperature is 1. Moving MNIST/FMNIST use $2\times2$ region pooling and occupancy from the last two observations with threshold 0.05, retaining the top 128 teacher references per anchor. Human3.6M selects the top 25\% of observed-motion locations; HMDB51 likewise uses motion-selected anchors. KittiCaltech enqueues 20 high-motion and 20 ordinary-region tokens while retaining all regions as anchors. KTH uses $16\times16$ forecast-difference regions, a 25\% motion fraction, and coordinates scaled by two.

BAIR stores at most one entry per episode, starts after 1,024 distinct episodes, and retrieves 64 references from other episodes. Its 384-dimensional key concatenates the latest RGB observation and signed first-to-last change after $8\times8$ pooling. WeatherBench Base configurations retain 64 historical entries per location and use latitude-weighted aggregation; the Tiny configuration instead uses a global FIFO with 1,024 sampled references. TaxiBJ stores 2,048 entries, retrieves 16 neighbors at temperature 0.1, and uses $4\times4$ pooling with horizon groups $\{1,2\}$ and $\{3,4\}$. All memory losses access history before the current detached teacher entries are inserted.

\clearpage
\section{Spatial Frequency Analysis Protocol}
\label{sec:spectral_protocol}

\paragraph{Sampling and predictions.}
For each of KTH and HMDB51, we select 256 windows at evenly spaced indices in the ordered test set before inspecting their predictions. The same windows are used for all models and four paired training runs (42--45), without filtering for positive gains. We evaluate every predicted frame: 20 for KTH and four for HMDB51. Both comparisons use U-Net-Base, with the checkpoint criterion described in Appendix~\ref{sec:experiment_protocols}. This is a descriptive analysis of fixed test subsets; overlapping windows are not treated as independent evidence of statistical significance.

\paragraph{Spectral transform.}
For a predicted or target frame channel $z\in\mathbb R^{H\times W}$, let $w$ denote a separable two-dimensional Hann window. We remove its weighted spatial mean $\mu_w(z)$ and compute an orthonormal two-dimensional Fourier transform:
\begin{equation}
    F(z)=\operatorname{FFT}_2\!\left(\frac{w\odot(z-\mu_w(z))}{\sqrt{\langle w^2\rangle}}\right),
    \qquad
    \mu_w(z)=\frac{\sum_{u,v}w_{uv}z_{uv}}{\sum_{u,v}w_{uv}}.
    \label{eq:spectral_transform}
\end{equation}
Here $\langle w^2\rangle$ is the spatial mean squared window value. We use raw floating-point predictions without clipping, sharpening, or contrast adjustment. DC coefficients are excluded. Consequently, the spectral analysis omits per-frame mean-brightness discrepancies and is not identical to pixel MSE. Power is averaged over native image channels and all prediction times.

\paragraph{Radial spectra and band errors.}
Radial power averages $|F(z)|^2$ within 48 uniformly spaced frequency annuli. Radial spatial frequency is measured in cycles per pixel; its maximum is $\sqrt{1/2}$ at the corners of the two-dimensional frequency grid. Low, middle, and high bands are $(0,1/8]$, $(1/8,1/4]$, and $(1/4,\sqrt{1/2}]$, respectively. For model $m$, training run $s$, and band $b$, we compute
\begin{equation}
    E_{m,s}^{(b)}=
    \frac{\sum_{i,t,c}\sum_{\boldsymbol\omega\in\Omega_b}
    |F(\widehat Y_{m,s}^{i,t,c})(\boldsymbol\omega)-F(Y^{i,t,c})(\boldsymbol\omega)|^2}
    {\sum_{i,t,c}\sum_{\boldsymbol\omega\in\Omega_b}
    |F(Y^{i,t,c})(\boldsymbol\omega)|^2}.
    \label{eq:spectral_error}
\end{equation}
The indices $i,t,c$ cover sampled windows, future times, and channels. This complex-valued residual retains phase discrepancies; greater power alone need not indicate more accurate detail. Each band is normalized by its own GT energy, so band heights do not measure their contributions to total pixel MSE. Figure~\ref{fig:spectral_analysis} reports means and population standard deviations across training runs. Relative reductions average $100(E_{\mathrm{S0},s}^{(b)}-E_{\mathrm{TAM},s}^{(b)})/E_{\mathrm{S0},s}^{(b)}$ over paired runs. Numerical checks verify Parseval energy conservation for the windowed fields.

\clearpage
\section{Motion and Temporal Analysis Protocol}
\label{sec:motion_protocol}

\paragraph{Test subsets and aggregation.}
We select 1,024 evenly spaced windows from each ordered test set before inspecting predictions. HMDB51 uses U-Net-Base and KittiCaltech uses ResNet-FCN, with the existing validation-selected checkpoints and four paired runs (42--45). Predictions are evaluated in floating point without clipping or post-processing. The sampled windows span 882 HMDB51 clips and 66 KittiCaltech recordings; overlapping windows are not treated as independent statistical replicates. For each run, MSE is averaged over windows after averaging the relevant pixels, channels, and future times within each window. Relative reductions are computed from paired run-level errors and then averaged across runs. Error bars use population standard deviations.

\paragraph{Input-defined motion groups and regions.}
For window $i$, we define a spatial change map and its mean strength as
\begin{equation}
    d_i(u,v)=\frac{1}{C}\sum_{c=1}^{C}\left|X^i_{T,c}(u,v)-X^i_{T-1,c}(u,v)\right|,
    \qquad a_i=\frac{1}{HW}\sum_{u,v}d_i(u,v).
    \label{eq:motion_proxy}
\end{equation}
Dataset-specific tertiles of $a_i$ define the low-, middle-, and high-motion groups. The high-change mask retains the top 20\% of pixels ranked by $d_i$, subject to $d_i>1/255$; stable pixel-index order breaks ties. Its complement defines the remaining region. Both masks are fixed from the input and shared across models, runs, and future frames. This proxy also captures camera and illumination changes and does not identify semantic foreground or track moving objects. Group analysis uses all 1,024 windows. Both regional metrics use the same windows with nonempty high-change masks: 983 for HMDB51 and 1,024 for KittiCaltech. Regional errors weight windows equally.

\begin{figure}[!htbp]
    \centering
    \includegraphics[width=0.65\linewidth]{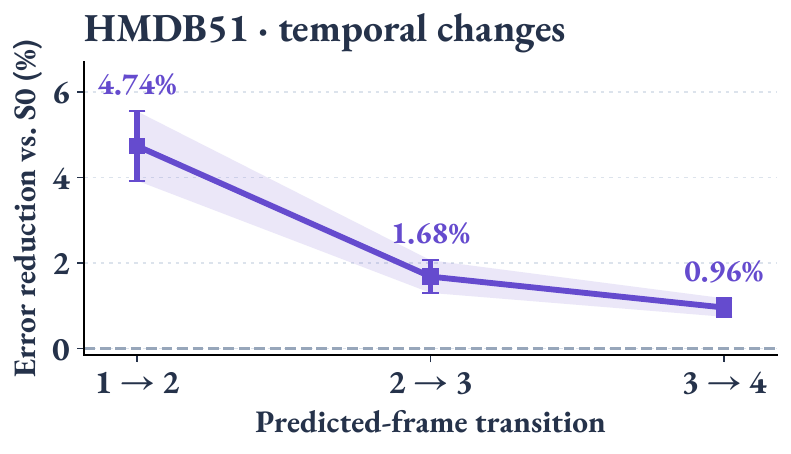}
    \caption{HMDB51 frame-difference MSE reduction relative to S0. Points and error bars show the mean and standard deviation over four paired runs.}
    \label{fig:temporal_changes}
\end{figure}

\paragraph{Reconstruction of temporal changes.}
For each transition between future frames, we compare predicted and target frame differences:
\begin{equation}
    D_{m,s,h}=\frac{1}{NCHW}\sum_{i=1}^{N}\left\|\bigl(\widehat Y^{i}_{m,s,h}-\widehat Y^{i}_{m,s,h-1}\bigr)-\bigl(Y^{i}_{h}-Y^{i}_{h-1}\bigr)\right\|_F^2,
    \quad h=2,3,4.
    \label{eq:temporal_change_error}
\end{equation}
Here $m$ and $s$ index the model and training run. On the same 1,024 HMDB51 windows, \method{} reduces this error relative to S0 by $4.74\%$, $1.68\%$, and $0.96\%$ across the three transitions (Figure~\ref{fig:temporal_changes}). Gains diminish at later transitions. The metric measures agreement with observed temporal changes rather than visual smoothness alone. KittiCaltech predicts one future frame and therefore does not support this within-forecast analysis.

\end{document}